\pdfoutput=1

\documentclass[11pt]{article}

\usepackage[final]{latex/acl}

\usepackage{times}
\usepackage{latexsym}

\usepackage[T1]{fontenc}
\usepackage[table,xcdraw]{xcolor}
\usepackage{pifont}
\usepackage[utf8]{inputenc}

\usepackage{microtype}

\usepackage{inconsolata}
\usepackage{listings}
\usepackage{arydshln}
\usepackage{graphicx}
\usepackage{times}
\usepackage{epsfig}
\usepackage{booktabs}
\usepackage{amsmath}
\usepackage{amssymb}
\usepackage{adjustbox}
\usepackage{multirow}
\usepackage{multicol}
\usepackage{caption}
\usepackage{color}
\usepackage{graphics}
\usepackage{algpseudocode}
\usepackage{algorithm}
\usepackage{tabularx}
\usepackage{cellspace}
\usepackage{float}
\usepackage{lipsum}
\usepackage{stfloats}
\usepackage{subcaption}
\usepackage{placeins}

\lstdefinestyle{PromptStyle}{
  basicstyle=\ttfamily\small,
  keywordstyle=\bfseries\color{blue},
  commentstyle=\color{gray},
  morekeywords={\{, \}},
  frame=single,
  breaklines=true
}

\title{Open Your Model's Eyes:\\Video and Context-Aware Multimodal Backchannel Prediction}

\author{
 \textbf{Min-Jae Kim\textsuperscript{1,*}},
 \textbf{Jun-Yeong Moon\textsuperscript{1,*}},
 \textbf{Mujeen Sung\textsuperscript{2,\textdagger}},
 \textbf{Gyeong-Moon Park\textsuperscript{1,\textdagger}}
\\
 \textsuperscript{1}Korea University, Seoul, Republic of Korea \\
 \textsuperscript{2}Kyung Hee University, Yongin, Republic of Korea
\\
 \small{\texttt{
   {\{minjaekim, moonjunyyy, gm-park\}@korea.ac.kr, mujeensung@khu.ac.kr}}
 }
}

\begin{document}
\maketitle
\renewcommand{\thefootnote}{\fnsymbol{footnote}}
\footnotetext[1]{Equal contribution.}
\footnotetext[2]{Corresponding author.}
\begin{abstract}
Backchannels, which signal listener states like empathy and understanding, are fundamental to natural human interaction.
However, current approaches rely solely on audio and text. This omits crucial visual cues, such as facial expressions and gestures, as well as broader conversational contexts, which are necessary for accurate prediction.
In this paper, we introduce Context-Aware Multimodal Alignment for Backchannel Prediction (CAMA-BC), a novel framework that leverages visual information through Multi-Layer Multimodal Alignment (MMA).
Our alignment process comprises two stages. First, Context Alignment (MMA-CA) utilizes unlabeled dialogues with videos to capture conversational contexts.
Next, Backchannel Alignment (MMA-BA) fine-tunes the representations specifically for backchannel prediction.
Experimental results show that CAMA-BC significantly outperforms both existing methods and simple multimodal baselines, with particular effectiveness in recognizing complex backchannels such as empathy.
\end{abstract}

\section{Introduction}
Human conversation is a dynamic interplay involving continuous signal exchange to co-construct meaning and maintain mutual understanding.
Central to this interaction is \textbf{backchanneling}: the production of brief yet essential conversational responses that provide real-time feedback on comprehension, engagement, and affective states of interlocutors \citep{Yngve70}.
From the initial effort that used simple linear classification \cite{kawahara16b_interspeech}, backchannel prediction has emerged as a critical component in natural language processing (NLP) \citep{ortega2020oh, jang-etal-2021-bpm, ortega2023modeling}.

Existing backchannel prediction methods have relied exclusively on linguistic information, such as textual and acoustic signals.
Those approaches systematically overlook the visual modality, which is essential for improving comprehension and facilitating mutual understanding in human interactions \cite{krauss1996nonverbal, scherer2013functions, mandal2014nonverbal} because of its richness in facial expressions, gestures, and pre-utterance movements.
Our preliminary investigation in Table \ref{tab:video_f1} reveals that incorporating video features from VideoMAE \cite{tong2022videomae} through simple concatenation yields only modest improvements.
While this confirms that visual information contains valuable signals for backchannel prediction, it simultaneously reveals a critical paradox: naive integration approaches fail to unlock the potential of visual modality.

\begin{figure}
    \centering
    \includegraphics[width=\linewidth]{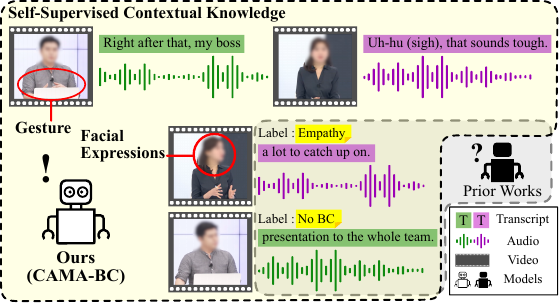}
    \caption{Schematic of the proposed data utilization in Context-Aware Multimodal Alignment for Backchannel Prediction (\textbf{CAMA-BC}), illustrating its ability to leverage a significantly broader range of information compared to the constrained scope of existing methods. }
    \label{fig:enter-label}
\end{figure}

This suggests a fundamental challenge in multimodal backchannel prediction: informational asymmetry across modalities.
Unlike text-based NLP tasks, where information flows sequentially, backchanneling requires understanding the subtle shift of information dominance between modalities over time.
Traditional encoder-concatenation architectures, which treat all modalities equally, fail to account for temporal offsets between modalities, thereby contributing to and ignoring changes in information dominance.
This asymmetry manifests in different contribution patterns: Audio contains both prosodic reactive cues and temporal contextual information, text provides rich semantic context but limited reactive signals, and video offers crucial non-verbal reactive cues.

Moreover, existing methods of backchannel prediction suffer from an imbalance between context and reaction.
As demonstrated in Table \ref{tab:dialog_quant}, backchannel events represent only a sparse subset of conversational data.
Additionally, the inherent subjectivity and semantic complexity of backchannel categories, such as empathy, exacerbate the training objective.
These challenges cannot be resolved with simple perturbations, such as data augmentation or class re-weighting.
This leads the models to overfit toward immediate pre-backchannel triggers, while neglecting broader conversational dynamics that inform natural backchanneling behavior.
As a result, models are limited to pattern-matching immediate signals rather than developing genuine conversational understanding.

\begin{table}[t]
\centering
  \resizebox{\linewidth}{!}{
  \begin{tabular}{Sc|ScSc}
      \specialrule{1.2pt}{-1.2pt}{0pt}
                        & w/o video  & w/ video \\ \hline
      BPM\_MT \cite{jang-etal-2021-bpm}    & 47.97 & \textbf{53.77}\\
      KoBERT+HuBERT         & 55.82 & \textbf{56.68}\\
      \specialrule{1.2pt}{-1.2pt}{0pt}
  \end{tabular}}
  \vspace{-2mm}
  \caption{Effect of the video modality adoption on F1 score in backchannel prediction on KC-Dialog.}
  \label{tab:video_f1}
  \vspace{-2mm}
\end{table}
\begin{table}[t]
  \centering
  \resizebox{\linewidth}{!}{
    \begin{tabular}{Sc|ScSc}
      \specialrule{1.2pt}{-1.2pt}{0pt}
                      & Text (\# words) & Audio \& Video (sec) \\
      \hline
        BC Sample  & 257,621             & 121,398\\
        Whole Dialog & 528,995             & 195,295\\
      \hline
      \hline
        Ratio (\%)   & \textbf{48.70}      & \textbf{62.16}\\
      \specialrule{1.2pt}{-1.2pt}{0pt}
  \end{tabular}}
  \vspace{-2mm}
  \caption{Comparison of time duration and word count between backchannel and whole dialogue datasets. BC Sample includes both NoBC and BC samples used for backchannel alignment.}
  \label{tab:dialog_quant}
  \vspace{-4mm}
\end{table}

To address these challenges of visual information integration, informational asymmetry, and context-reaction imbalance, we introduce \textbf{CAMA-BC (Context-Aware Multimodal Alignment for Backchannel Prediction)}. This novel framework learns robust representations from unlabeled full dialogues while modeling context-dependent modality dominance through adaptive cross-attention.
In this work, we propose a \textbf{Multi-Layer Multimodal Alignment (MMA)}, which utilizes hierarchical cross-attention mechanisms that account for information density differences across modalities.

Recognizing that backchannel prediction requires contextual depth, which cannot be acquired from sparse, noisy annotations alone, we employ a curriculum learning approach \cite{ELMAN199371} consisting of two stages.
First, \textbf{Context Alignment (MMA-CA)} is an unsupervised stage where the model learns rich conversational dynamics from entire dialogues.
This stage enables the model to understand broad conversational patterns, acting as a regularization that prevents the upcoming stage from being biased towards reactive triggers.
Next, \textbf{Backchannel Alignment (MMA-BA)} is a supervised stage that specializes in learned contextual representations for precise backchannel prediction.
This stage maintains contextual understanding while developing task-specific reactive precision through cross-attention-enhanced features.
Our framework offers a comprehensive solution that synergistically integrates contextual understanding with reactive precision, bridging the gap between conversational comprehension and backchannel prediction in multimodal settings.

Our main contributions can be summarized as follows:
\begin{itemize}
    \item We identify the crucial role of the visual modality in capturing reactive components for backchannel prediction and propose a scheme that effectively integrates video information.
    \item To model context-dependent modality dominance, we propose \textbf{Multi-Layer Multimodal Alignment (MMA)}, which aligns multimodal features across multiple layers through adaptive cross-attention with shared weights. 
    \item We propose a novel curriculum learning approach that consists of \textbf{Context Alignment (MMA-CA)} and \textbf{Backchannel Alignment (MMA-BA)}. The former leverages whole conversations to acquire rich contextual components, while the latter focuses on backchannel-specific information while maintaining contextual understanding.
    \item Through extensive experiments, we demonstrate that \textbf{Context-Aware Multimodal Alignment for Backchannel Prediction (CAMA-BC}) provides a practical and robust solution for backchannel prediction.
\end{itemize}

\section{Related Work}
\subsection{Backchannel Prediction}

Typical text-based natural language processing (NLP) models have shown success in various tasks such as speech recognition \cite{44926, xu2021self, shakhadri2025samba}, machine translation \cite{sennrich2016linguistic, weifinetuned}, sentiment analysis \cite{lan2019albert, jiang2020smart, raffel2020exploring}, and question-answering \cite{zhong2022toward, chowdhery2023palm}.

Since \citet{kawahara16b_interspeech} formalized backchannel prediction as a computational task, researchers have recognized backchannels as crucial signals for conveying empathy and agreement in dialogue \citep{Yngve70}.
Early approaches, such as \citet{ortega2020oh}, demonstrated that incorporating transcribed text and audio can significantly improve backchannel prediction.
Furthermore, they utilized listener identity information to enhance prediction.
Further research on BPM\_MT \cite{jang-etal-2021-bpm} aimed to increase accuracy by employing larger models and integrating sentiment classification as an auxiliary task, providing additional context that enhanced the predictive capabilities.
Recent work by \citet{ortega2023modeling} extended this line by incorporating listener and speaker identities, emphasizing the speaker-listener interaction.

Previous studies have predominantly relied on audio-text combinations, systematically excluding non-verbal signals, such as visual information, despite its documented importance in human communication \cite{krauss1996nonverbal, scherer2013functions}.
Additionally, simple feature concatenation fails to account for the differential temporal dynamics of linguistic and non-linguistic signals, treating all modalities as if they contribute equally at every timestep.
This narrow focus often neglects the broader conversational context that informs natural backchanneling behavior.
Our approach addresses these limitations through context-aware, multimodal alignment, which captures both long-term conversational dynamics and immediate reactive cues across text, audio, and visual modalities.
\begin{table*}[t!]
\centering
\resizebox{0.9\textwidth}{!}{%
\begin{tabular}{ccc}
\specialrule{1.2pt}{-1.2pt}{0pt}
Backchannel Category & Description & \multicolumn{1}{c}{Examples} \\ \hline
NoBC & No backchannel signal. & - \\
Continuer & Short or repeated sound indicating active listening. & "neh", "yeh", "ah" \\
Understanding & Longer cues showing understanding. & "uhm-", "uh-", "ah-" \\
Empathy & Emotions of the listener, such as surprise, sympathy, or disappointment. & "hah", "ugh", "whoa" \\ 
\specialrule{1.2pt}{-1.2pt}{0pt}
\end{tabular}%
}
\vspace{-2mm}
\caption{Descriptions and examples of the Backchannel categories used in this work.}
\vspace{-5mm}
\label{tab:bc_label}
\end{table*}
\vspace{-2mm}
\subsection{Multimodal Alignment}
Multimodal alignment seeks to integrate heterogeneous data sources, including text, audio, and video, into a unified representation space, thereby enabling a more nuanced understanding of semantics and facilitating cross-modal interactions.
As interest in multimodal research continues to rise, remarkable progress \cite{wang-etal-2020-maf, rouditchenko2021avlnet, sun, praveen2022joint, 10.1145/3503161.3548009, 9495252, shvetsova2022everything, Wang_2022_CVPR, sadoughi2023mega, girdhar2023imagebind, he2023align, zhou2024token, zhu2024languagebind} has emerged.
However, many works still struggle to align more than two modalities, failing to adequately handle the geometric increase in complexity that accompanies the combination of three or more modalities.

\citet{rouditchenko2021avlnet} proposed a self-supervised framework that aligns audio and raw video inputs in a joint embedding space without text annotations.
\citet{shvetsova2022everything} proposed a multimodal fusion transformer robust to the modalities and lengths, using combinatorial contrastive loss. 
ImageBind \cite{girdhar2023imagebind} focused on images, and LanguageBind \cite{zhu2024languagebind} centered on language, exploring efficiency by aligning one modality as an anchor. 
\citet{vaswani2017attention} facilitated alignment using cross-attention mechanisms, where the modality with dominant performance serves as both the key and value.

These methods have demonstrated remarkable advances; however, conversational alignment remains underexplored. 
Unlike static contexts, conversations involve dynamically varying temporal cues and uneven information density, posing unique alignment challenges.
In our approach, we depart from the static alignment assumption, ensuring alignment between each modality and its combinations.
We enforce weight-sharing across the higher encoder layers to decouple the roles of encoding and cross-modal alignment, thereby encoding consistent spatiotemporal dependencies across modalities.
Additionally, a two-stage curriculum learning framework is proposed that first captures broad conversational patterns before specializing in backchannel prediction.

\begin{figure*}[h!]
    \centering
      \includegraphics[width=1.0\textwidth]{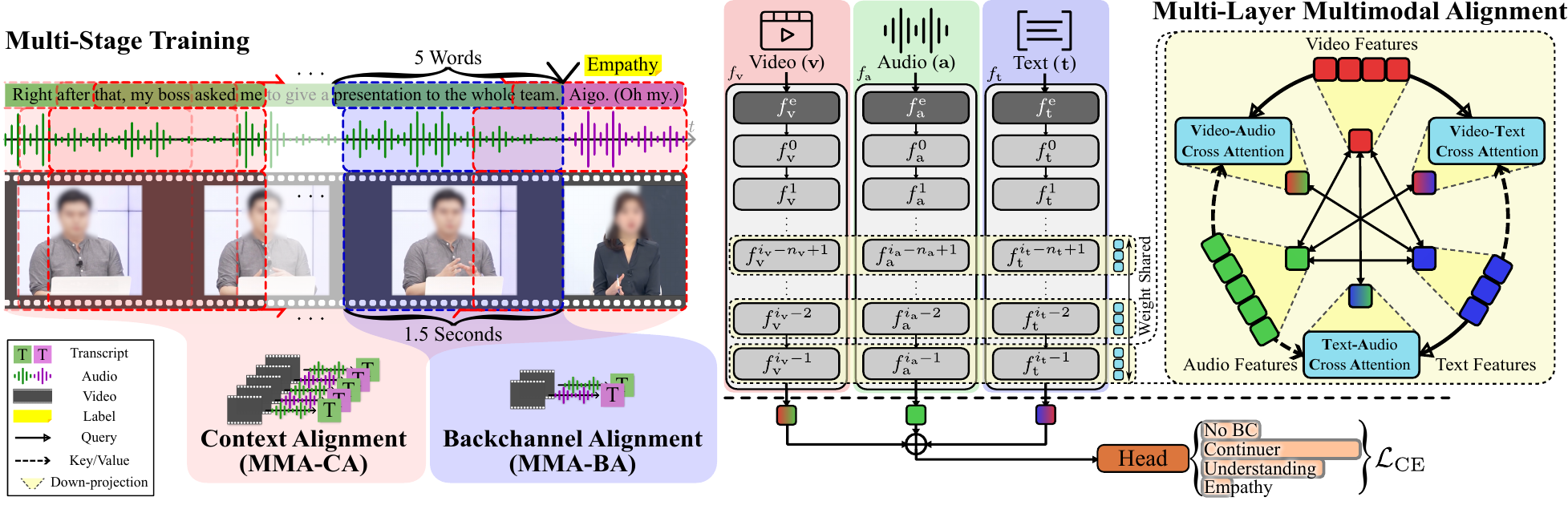}
    \caption{Illustration of the data construction and architecture of Context-Aware Multimodal Alignment for Backchannel Prediction (CAMA-BC), highlighting the Multi-Layer Multimodal Alignment (MMA) framework. The data construction process demonstrates the broader scope of the Context Alignment (MMA-CA).}
    \label{fig:mainfig}
\end{figure*}

\section{Method}
\subsection{Problem Formulation}

Given a conversation, we consider three synchronous modalities: audio, transcript, and newly considered video data. The audio dataset $\mathcal{D}_\mathrm{a}$ consists of amplitude values sampled at rate $s_\mathrm{a}$, and the transcript dataset $\mathcal{D}_\mathrm{t}$ is an ordered set of the words with their corresponding timestamps as follows:
\begin{align}
    \mathcal{D}_\mathrm{a} &= \left\{a_i \mid 0 \leq i < s_\mathrm{a} l_d \right\}, \label{eq:audio_data}\\
    \mathcal{D}_\mathrm{t} &= \left\{(t_i, p_i) \mid 0 \leq i < n_\mathrm{t} \right\}, \label{eq:text_data}
\end{align}
where $a_i \in \mathbb{R}$ represents audio amplitude, $t_i \in \mathbb{R}$ indicates the timestamp of each word, $p_i \in \mathcal{V}$ denotes a word from vocabulary $\mathcal{V}$, $l_d$ is the conversation duration, and $n_t$ is the total number of words.
For the first time, in this work, we introduce video data to the backchannel prediction task, enabling the model to capture non-verbal cues in dialogue:
\begin{align}
    \mathcal{D}_\mathrm{v} &= \left\{\mathbf{v}_i \mid 0 \leq i < s_\mathrm{v} l_d \right\}, \label{eq:video_data}
\end{align}
where $\mathbf{v}_i \in \mathbb{R}^{c \times w \times h}$ represents a video frame with $c$ channels and spatial dimensions $w \times h$, sampled at frame rate $s_v$.

The dataset includes annotations for listener engagement events, where each event is characterized by its timestamp $t^\mathrm{BC}_i \in \mathbb{R}$, vocabulary $p^\mathrm{BC}_i \in \mathcal{V}$ of the transcribed conversation, and category $y_i$ from $k$ possible backchannel classes $\mathcal{Y} = \{y_1, \ldots, y_k\}$:
\begin{align}
    (t^\mathrm{BC}_i, p^\mathrm{BC}_i, y_i) \in \mathcal{D}_\mathrm{BC}, \label{eq:bc_data}
\end{align}
where $ 0 < i \leq \left| \mathcal{D}_\mathrm{BC}  \right| $.
The detailed meaning of each category is demonstrated in Table \ref{tab:bc_label}.
For each backchannel event $(t_i^\mathrm{BC}, p_i^\mathrm{BC}, y_i) \in \mathcal{D}_\mathrm{BC}$, we extract context from the preceding $n$ seconds for the audio and video, and $m$ words for the text:
\begin{align}
    \mathcal{W}_{\mathrm{a}, i}^\mathrm{BC} &= \left\{a_j \mid \left(t_i^\mathrm{BC} - n \right)s_a \leq j < t_i^\mathrm{BC}s_a \right\}, \label{eq:audio_window}\\
    \mathcal{W}_{\mathrm{t}, i}^\mathrm{BC} &= \left\{p_j \mid i - m \leq j < i \right\},\label{eq:text_window}\\
    \mathcal{W}_{\mathrm{v}, i}^\mathrm{BC} &= \left\{v_{\sigma(k)} \mid k \in \{1, 2,\ldots,l\}\right\}, \label{eq:video_window}
\end{align}
where $\sigma(k) = \left\lfloor(t^{\mathrm{BC}}_i - n)s_v + (k-1)\frac{ns_v}{l-1}\right\rfloor$ uniformly samples $l$ frames over the interval $[t_i^{BC}-n,\; t_i^{BC})$, excluding the frame at the backchannel onset $t_i^{BC}$.
The extracted windows are then processed into fixed-size tensors:
\begin{align}
    \mathbf{a}_{i} &= \phi(\mathcal{W}_{\mathrm{a}, i}^\mathrm{BC}) \in \mathbb{R}^{b \times ns_a}, \label{eq:audio_tensor}\\
    \mathbf{t}_{i} &= \tau(\mathcal{W}_{\mathrm{t}, i}^\mathrm{BC}) \in \mathbb{N}^{b \times m_\mathrm{pad}}, \label{eq:text_tensor}\\
    \mathbf{v}_{i} &= \phi(\mathcal{W}_{\mathrm{v}, i}^\mathrm{BC}) \in \mathbb{R}^{b \times l \times c \times w \times h}, \label{eq:video_tensor}
\end{align}
where $\phi(\cdot)$ stacks tensors from a set, $\tau(\cdot)$ tokenizes text to integers with padding length $m_\mathrm{pad}$, and $b$ denotes the batch size.
The final training batch $\mathcal{B}$, also called backchannel alignment dataset, combines these tensors:
\begin{align}
    \mathcal{B}_i &= \{\mathbf{a}_i, \mathbf{t}_i, \mathbf{v}_i, y_i\}.
    \label{eq:batch}
\end{align}
In practical implementation, it provides only the utterances of one speaker, with the utterances of the other speaker masked, for a realistic task configuration.
The masking strategy is used as a preprocessing step to reduce input ambiguity caused by speaker alternation and overlap, rather than to encode speaker identity. 
We do not enforce a speaker-disjoint split, because the task focuses on modeling listener reactions rather than speaker behavior.

\subsection{Context Alignment Dataset} \label{sec:ctx_data}
As illustrated in Figure \ref{fig:enter-label} and Table \ref{tab:dialog_quant}, backchannel responses constitute only a subset of the total conversational data, with a severe class imbalance for semantic categories such as empathy.
This sparsity creates two fundamental challenges: (1) supervised learning on backchannel-only data provides insufficient examples for acquiring a robust pattern of the conversation, (2) models trained exclusively on pre-backchannel segments exhibit reactive bias, focusing solely on immediate triggers rather than understanding broader conversational flow.

To address these limitations, we propose Context Alignment (CA), an unsupervised pre-training phase that leverages the complete conversational corpus.
For each word-timestamp pair $(t_i^\mathrm{CA}, p_i^\mathrm{CA}) \in \mathcal{D}_\mathrm{CA}$, we extract context windows of identical dimensions to those used in backchannel prediction.
By maintaining identical window dimensions across two phases, we ensure that contextual representations learned during unsupervised pre-training serve as initialization for fine-tuning the model towards backchannel prediction.
In contrast, the model learns to distinguish between general conversational patterns and backchannel-specific triggers.
By exposing the model to temporally synchronized multimodal context beyond backchannel events, the pre-training encourages temporal alignment across modalities.

\begin{table}[t]
\centering
\resizebox{\linewidth}{!}{%
\begin{tabular}{ccccccc}
\specialrule{1.2pt}{-1.2pt}{0pt}
\multirow{2}{*}{Model} & \multirow{2}{*}{NoBC} & \multicolumn{3}{c}{BC} & \multirow{2}{*}{Macro F1} \\ \cline{3-5}
 &  & Continuer & Understanding  & Empathy &  \\ \hline
HuBERT (audio) & 85.21\tiny{$\pm$ 0.31} & 65.77\tiny{$\pm$ 1.40} & 49.14\tiny{$\pm$ 0.46} & 17.05\tiny{$\pm$ 1.11} & 54.29\tiny{$\pm$ 0.11}  \\
KoBERT (text) & 84.09\tiny{$\pm$0.74} & 57.90\tiny{$\pm$1.94} & 30.83\tiny{$\pm$0.65} & 17.45\tiny{$\pm$ 1.07} & 47.57 \tiny{$\pm$0.66} \\
VideoMAE (video) & 74.91\tiny{$\pm$ 0.32} & 53.98\tiny{$\pm$ 1.10} & 35.16\tiny{$\pm$ 0.74} & 0.00\tiny{$\pm$ 0.00} & 41.01\tiny{$\pm$ 0.09} \\ 
\specialrule{1.2pt}{-1.2pt}{0pt}
\end{tabular}
}
\vspace{-2mm}
\caption{F1 scores on KC-Dialog dataset with a single-modality encoder, to evaluate the impact of each modality on the backchannel prediction task.}
\label{tab:single_modal}
\vspace{-3mm}
\end{table}

\subsection{Multi-Layer Multimodal Alignment}
\label{sec:mma}
Unlike the backchannel alignment dataset, the context data constructed in Section \ref{sec:ctx_data} does not contain explicit labels.
Therefore, we need an approach to learning contextual information from unlabeled dialogue data.
Our preliminary analysis (Table \ref{tab:single_modal}) reveals asymmetric information density across modalities for backchannel prediction.
Audio shows the highest predictive power, containing both reactive and contextual information.
The text is followed by providing rich contextual semantics, but with limited reactive cues.
In contrast, video contains sparse linguistic content but contributes crucial non-verbal reactive signals.

This asymmetry motivates a hierarchical alignment strategy where information-dense modalities assist information-sparse ones, rather than treating all modalities equivalently.
It determines explicitly which modality acts as the Query and which as the Key/Value in cross-modal interactions, prioritizing them in the order of audio, text, and video through cross-modal feature alignment.

As shown in Figure \ref{fig:mainfig}, we distinguish between embedding layers ($f_\mathrm{a}^e$, $f_\mathrm{t}^e$, and $f_\mathrm{v}^e$) and encoder layers ($\{f_\mathrm{a}^l ~ | ~ 0\leq l<i_\mathrm{a}\}$, $\{f_\mathrm{t}^l ~| ~ 0\leq l<i_\mathrm{t}\}$, and $\{f_\mathrm{v}^l ~| ~0\leq l<i_\mathrm{v}\}$) for audio, text, and video models, respectively.
The feature extraction process can be expressed as:
\begin{align}
    \mathbf{a}^0_i &= f_\mathrm{a}^e(\mathbf{a}_i),\;\, \mathbf{a}^{l+1}_i = f_\mathrm{a}^l(\mathbf{a}^{l}_i),\nonumber\\
    \mathbf{t}^0_i &= f_\mathrm{t}^e(\mathbf{t}_i),\;\; \mathbf{t}^{l+1}_i = f_\mathrm{t}^l(\mathbf{t}^{l}_i),\nonumber\\
    \mathbf{v}^0_i &= f_\mathrm{v}^e(\mathbf{v}_i),\; \mathbf{v}^{l+1}_i = f_\mathrm{v}^l(\mathbf{v}^{l}_i).
\end{align}
We align the features in the last $n_l$ layers of each modality encoder as follows:
\begin{align}
    \mathcal{J} &= \left\{
        j=\left(j_\mathrm{a}, j_\mathrm{t}, j_\mathrm{v}\right) \mid
        \right.\nonumber\\&\qquad\qquad\left.
        i_\mathrm{a}-n_\mathrm{a}+1 \leq j_\mathrm{a} < i_\mathrm{a},
        \right.\nonumber\\&\qquad\qquad\left.
        i_\mathrm{t}-n_\mathrm{t}\;+1 \leq j_\mathrm{t} < i_\mathrm{t},
        \right.\nonumber\\&\qquad\qquad\left.
        i_\mathrm{v}-n_\mathrm{v}+1 \leq j_\mathrm{v} < i_\mathrm{v}\right\},
\end{align}
for audio, text, and video, respectively, with the same number of selections $n_\mathrm{a}=n_\mathrm{t}=n_\mathrm{v}=n_l$.
This strategic choice emphasizes semantic alignment rather than modality-specific features.
For the alignment at the $j = \left(j_\mathrm{a}, j_\mathrm{t}, j_\mathrm{v}\right)$, we first project the sequence-level features into a shared space and then average-pool the projected features into sample-level embeddings.
The model aligns embeddings from the same sample while treating the remaining samples in the batch as negatives:





\begin{align}
\mathcal{L}_\mathrm{MMA}^j &= \sum_{\substack{\alpha\in\mathcal{S}_j,\beta\in\mathcal{S}_j \setminus \left\{\alpha\right\}}}\delta(\alpha,\beta), \label{eq:MMA} \\
\delta(\mathbf{v},\mathbf{w})
&=
-\operatorname{diag}\!\left(
\operatorname{LogSoftmax}\!\left(
\frac{\mathbf{v}^{\top}\mathbf{w}}
{\|\mathbf{v}\|\,\|\mathbf{w}\|}
\right)
\right), \nonumber \\
\mathcal{S}_j
&=
\left\{
\bar{\mathbf{a}}^{j_\mathrm{a}},
\bar{\mathbf{t}}^{j_\mathrm{t}},
\bar{\mathbf{v}}^{j_\mathrm{v}}
\right\}, \label{eq:set_x} 
\end{align}
where $\bar{\mathbf{m}}^{j_\mathrm{m}} = \operatorname{AvgPool}\!\left(W_\mathrm{m}\mathbf{m}^{j_\mathrm{m}}\right)$ denotes the sample-level embedding of modality $\mathrm{m} \in \{\mathrm{a}, \mathrm{t}, \mathrm{v}\}$, and $W_\mathrm{a} \in \mathbb{R}^{d_\mathrm{a}\times d}$, $W_\mathrm{t} \in \mathbb{R}^{d_\mathrm{t}\times d}$, and $W_\mathrm{v} \in \mathbb{R}^{d_\mathrm{v}\times d}$ are linear transformations that project encoder features into a shared $d$-dimensional space.
Our empirical analysis shows that applying alignment to earlier layers proves counterproductive.
These layers lack the necessary abstraction for effective cross-modal fusion, primarily capturing modality-specific features.

However, the representations across video, audio, and text modalities exhibit significant differences, making direct alignment using simple alignment difficult.
To address this issue, we employ a hierarchical cross-attention mechanism, motivated by our experimental findings indicating varying information densities (e.g., audio > text > video) and distributions across modalities, as shown in Table \ref{tab:single_modal}.
The cross-attention layer enables a query modality to be refined by selectively attending to information from an assistant key/value modality with denser information.
To incorporate attention-guided enhancement, we extend the alignment loss with average-pooled cross-attended representations as follows:
\begin{align}
    \mathcal{L}^{j,*}_\mathrm{MMA} &= \mathcal{L}_\mathrm{MMA}^j
    + \delta\left(\bar{\mathbf{a}}^{j_\mathrm{a}}, \bar{\mathbf{c}}_{\mathrm{vt}}^{\,j}\right)\nonumber\\  
    &+ \delta\left(\bar{\mathbf{t}}^{j_\mathrm{t}}, \bar{\mathbf{c}}_{\mathrm{va}}^{\,j}\right) 
    + \delta\left(\bar{\mathbf{v}}^{j_\mathrm{v}}, \bar{\mathbf{c}}_{\mathrm{ta}}^{\,j}\right),
    \label{eq:CAA}
\end{align}


where $\bar{\mathbf{c}}_{\mathrm{vt}}^{\,j}$, $\bar{\mathbf{c}}_{\mathrm{va}}^{\,j}$, and $\bar{\mathbf{c}}_{\mathrm{ta}}^{\,j}$ denote the average-pooled cross-attended representations computed from video-text, video-audio, and text-audio feature pairs, respectively.
The hierarchy ensures that information flows from denser key/value modalities to sparser query modalities. 
For instance, in $\delta\left(\bar{\mathbf{a}}^{j_\mathrm{a}}, \bar{\mathbf{c}}_{\mathrm{vt}}^{\,j}\right)$, the video feature $\bar{\mathbf{v}}^{j_\mathrm{v}}$ is used as the query and the relatively informative text feature $\bar{\mathbf{t}}^{j_\mathrm{t}}$ is used as the key and value, producing a representation aligned with the audio feature $\bar{\mathbf{a}}^{j_\mathrm{a}}$.

Traditional layer-specific attention would learn modality-specific transformations at each layer, potentially leading to overfitting to layer-dependent features and competing for the same knowledge space as the encoders.
Instead, we share cross-attention weights across all layers in $\mathcal{J}$.
This strategy encourages the cross-attention layer to learn consistent cross-modal interaction patterns and a modality-agnostic approach to identifying salient inter-token relationships.
This approach facilitates smoother learning by robustly pinpointing crucial tokens across different representational levels.

This learned attention pattern is then directly utilized in the final prediction stage as follows:
\begin{align}
    \mathbf{y}_{\mathrm{pred}}
    =
    W_h \cdot \mathrm{cat}\left\{
        \bar{\mathbf{a}}^{j_\mathrm{a}},
        \bar{\mathbf{c}}_{\mathrm{va}}^{\,j},
        \bar{\mathbf{c}}_{\mathrm{ta}}^{\,j}
    \right\} + \mathbf{b}_h,
\end{align}
where $\mathbf{y}_{\mathrm{pred}}$ denotes the logits, $\mathrm{cat}\left(\cdot, \cdot, \dots \right)$ is concatenation of the given tensors, and $W_h\in\mathbb{R}^{3d\times k}$ and $\mathbf{b}_h\in\mathbb{R}^{k}$ are weight and bias in the classification head, respectively.
While the audio serves as the primary anchor, video and text are enhanced using dense information from the audio.
This architecture preserves the information hierarchy while enabling cross-modal enhancement.

\begin{table*}[t!]
\centering
\resizebox{1.\textwidth}{!}{%
\begin{tabular}{ccccc||cccc||c}
\specialrule{1.2pt}{-1.2pt}{0pt}
\multirow{2}{*}{Dataset} & \multirow{2}{*}{NoBC} & \multicolumn{3}{c||}{BC} & \multirow{2}{*}{Train} & \multirow{2}{*}{Validation} & \multirow{2}{*}{Test} & \multirow{2}{*}{Total} & \multirow{2}{*}{\begin{tabular}{c} \#Samples \\ for MMA-CA \end{tabular}} \\ \cline{3-5}
 & & Continuer & Understanding  & Empathy & & & & & \\ \hline
KC-Dialog & 34,710 & 19,635 & 9,494 & 5,581 & 55,536 & 3,470 & 10,414 & 69,420 & 130,278 \\
BACKSpeech & 4,907 & 4,046 & 352 & 509 & 7,852 & 490 & 1,472 & 9,814 & 34,580 \\
\specialrule{1.2pt}{-1.2pt}{0pt}
\end{tabular}%
}
\vspace{-2mm}
\caption{Distribution of each backchannel category and the number of samples used for MMA-CA. Following prior works, we ensured that the NoBC and BC samples had equal numbers. The dataset was then split into training, validation, and test sets in an $ 8:0.5:1.5$ ratio.}
\label{tab:sample}
\vspace{-1mm}
\end{table*}
\subsection{Multi-Stage Training}
We construct our Multi-Layer Multimodal Alignment (MMA) framework through a two-phase training process: Context Alignment (MMA-CA) followed by Backchannel Alignment (MMA-BA).
The first phase (Section \ref{stage1_cont_align}) trains the model on general conversational context and cross-modal relationships.
In contrast, the second phase (Section \ref{stage2_back_align}) adapts the model specifically for backchannel prediction while preserving contextual knowledge.

\subsubsection{Context Alignment}
\label{stage1_cont_align}
The Context Alignment phase utilizes the data in Section \ref{sec:ctx_data} to train the model on broader conversational dynamics and inter-modal correlations. During this phase, we optimize solely using the MMA loss defined in Equation \ref{eq:CAA}:
\begin{align}
    \mathcal{L}_\mathrm{CA} = \sum_{j \in \mathcal{J}} \mathcal{L}^{j, *}_\mathrm{MMA},
\end{align}
where $\mathcal{J}$ represents the layers selected for alignment.
The model learns to recognize correlations between verbal, vocal, and visual elements of natural dialogue by training on continuous conversation segments of equivalent length to the backchannel-annotated samples.
This equivalence in segment length ensures that the learned representations can be directly applied to the downstream task.

\subsubsection{Backchannel Alignment}
\label{stage2_back_align}
The Backchannel Alignment phase fine-tunes the model on backchannel-annotated data while maintaining the contextual knowledge learned in the first phase.
We optimize a combined objective that incorporates both classification and alignment losses:
\begin{align}
    \mathcal{L}_\mathrm{BA} = \mathcal{L}_\mathrm{CE}\left(\mathbf{y}_{\mathrm{pred}}, y\right) + \lambda \sum_{j \in \mathcal{J}} \mathcal{L}^{j, *}_\mathrm{MMA},
\end{align}
where $\mathcal{L}_\mathrm{CE}\left(\cdot, \cdot\right)$ denotes the cross-entropy loss for backchannel classification, and $\lambda$ controls the contribution of the alignment loss (set to 0.1 in our experiments).
This dual objective serves two purposes: (1) it preserves contextual knowledge during task-specific adaptation, and (2) it strengthens the cross-modal relationships relevant to backchannel prediction.



\section{Experiments}

\subsection{Experimental Setup}
\noindent\textbf{Datasets.}
We utilize three datasets: the Korean Counseling Dialog Dataset (KC-Dialog), the Backchannel Annotation Corpus in Korean Speech Dataset (BACKSpeech) \cite{BACKSpeech}, and the Switchboard (SWBD) corpus \cite{godfrey1992switchboard}.
KC-Dialog consists of counselor-client videos, each about 50 minutes long, with transcripts annotated for backchannel classes.
BACKSpeech contains 55 videos totaling 20 hours from diverse media sources, including radio, news, and debates.
While both datasets provide backchannel-annotated transcripts, BACKSpeech features more dynamic visual content with varying scenes and visual materials, in contrast to speaker-centric recordings of KC-Dialog.
The audio is sampled at $s_a = 16000 \text{Hz}$ and video is sampled at $s_v = 30 \text{Hz}$. A description of SWBD is provided in Section~\ref{swbd}.

Following the categorization scheme of BPM\_MT \citep{jang-etal-2021-bpm}, we classify backchannels into four classes: NoBC, Continuer, Understanding, and Empathy, with examples and descriptions provided in Table \ref{tab:bc_label}, and class distributions detailed in Table \ref{tab:sample}.
For feature extraction, we analyze temporal windows of $n = 1500 \text{ms}$ and $m = 5$ words following each backchannel onset, adhering to established protocols from previous studies \cite{ortega2020oh, jang-etal-2021-bpm}.
Corresponding to the temporal segment of the video, we uniformly sample $l = 12$ frames following \citet{tong2022videomae}.

\begin{table*}[t]
\centering
\resizebox{0.9\textwidth}{!}{%
\begin{tabular}{cccccccc}
\specialrule{1.2pt}{-1.2pt}{0pt}
\multirow{2}{*}{Dataset} & \multirow{2}{*}{Model} & \multirow{2}{*}{Modality} & \multirow{2}{*}{NoBC} & \multicolumn{3}{c}{BC} & \multirow{2}{*}{Macro F1} \\ \cline{5-7}
& & & & Continuer & Understanding & Empathy & \\ \hline
\multirow{9}{*}{KC-Dialog} & KoBERT & T & 84.09\tiny{$\pm$0.74} & 57.90\tiny{$\pm$1.94} & 30.83\tiny{$\pm$0.65} & 17.45\tiny{$\pm$ 1.07} & 47.57 \tiny{$\pm$0.66}  \\ 
& Llama 4.0 Scout & T & 53.02\tiny{$\pm$0.61} & 25.07\tiny{$\pm$1.21} & 12.25\tiny{$\pm$0.86} & 4.48\tiny{$\pm$0.20} & 23.71\tiny{$\pm$0.20} \\
& GPT-4.1 & T & 57.39\tiny{$\pm$0.30} &	33.03\tiny{$\pm$0.47} &	9.86\tiny{$\pm$0.09} &	6.73\tiny{$\pm$0.06} &	26.75\tiny{$\pm$0.17} \\ \cdashline{2-8}
& Ortega & A,T & 84.53\tiny{$\pm$0.51} & 60.54\tiny{$\pm$1.20} & 24.25\tiny{$\pm$0.98} & 7.59\tiny{$\pm$ 1.33} & 44.23 \tiny{$\pm$0.83}  \\ 
& BPM\_MT & A,T & 85.73\tiny{$\pm$0.27} & 61.55\tiny{$\pm$1.72} & 30.43\tiny{$\pm$2.77} & 14.18\tiny{$\pm$ 2.33} & 47.97 \tiny{$\pm$0.78}  \\ \cdashline{2-8}
& Gemini-2.5-Flash & A,T,V & 44.63\tiny{$\pm$1.11} & 6.02\tiny{$\pm$1.24} & 3.62\tiny{$\pm$0.48} & 4.68\tiny{$\pm$0.89} & 14.74\tiny{$\pm$0.52} \\
& Ortega-V & A,T,V & 84.46\tiny{$\pm$0.39} & 65.16\tiny{$\pm$1.18} & 42.74\tiny{$\pm$1.00} & 7.25\tiny{$\pm$0.18} & 49.90\tiny{$\pm$0.27} \\ 
& BPM-V & A,T,V & 86.96\tiny{$\pm$0.18} & 65.72\tiny{$\pm$1.49} & 45.50\tiny{$\pm$1.97} & 16.89\tiny{$\pm$2.09} & 53.77\tiny{$\pm$0.09}  \\ 
\rowcolor{lightgray} & \textbf{CAMA-BC (Ours)} & A,T,V & \textbf{89.90}\tiny{$\pm$0.07} & \textbf{72.73}\tiny{$\pm$0.13} & \textbf{49.99}\tiny{$\pm$0.25} & \textbf{21.49}\tiny{$\pm$0.49} & \textbf{58.53}\tiny{$\pm$0.07}  \\ \hline
\multirow{9}{*}{BACKSpeech} & KoBERT & T & 67.04\tiny{$\pm$0.65} & 62.80\tiny{$\pm$0.21} & 0.00\tiny{$\pm$0.00} & 0.00\tiny{$\pm$ 0.00} & 32.46 \tiny{$\pm$0.21}  \\ 
& Llama 4.0 Scout & T & 18.69\tiny{$\pm$0.99} &	27.93\tiny{$\pm$0.26} &	4.63\tiny{$\pm$0.49} &	2.11\tiny{$\pm$0.39} &	13.34\tiny{$\pm$0.19} \\ 
& GPT-4.1 & T & 30.21\tiny{$\pm$0.39} & 43.73\tiny{$\pm$0.74} & 2.51\tiny{$\pm$0.36} & 3.12\tiny{$\pm$0.13} & 19.89\tiny{$\pm$0.33} \\ \cdashline{2-8}
& Ortega & A,T & 68.21\tiny{$\pm$0.60} & 61.39\tiny{$\pm$0.44} & 0.00\tiny{$\pm$0.00} & 0.00\tiny{$\pm$ 0.00} & 32.40 \tiny{$\pm$0.15}  \\ 
& BPM\_MT & A,T & 69.10\tiny{$\pm$1.03} & 62.66\tiny{$\pm$0.74} & 0.00\tiny{$\pm$0.00} & 0.00\tiny{$\pm$ 0.00} & 32.94 \tiny{$\pm$0.31}  \\ \cdashline{2-8}
& Gemini-2.5-Flash & A,T,V & 26.19\tiny{$\pm$2.42} & 11.64\tiny{$\pm$1.22} & 1.51\tiny{$\pm$1.16} & 3.65\tiny{$\pm$0.64} & 10.75\tiny{$\pm$0.39} \\
& Ortega-V & A,T,V & 69.24\tiny{$\pm$0.68} & 62.63\tiny{$\pm$0.33} & 0.00\tiny{$\pm$0.00} & 1.22\tiny{$\pm$ 1.06} & 33.27 \tiny{$\pm$0.02}  \\ 
& BPM-V & A,T,V & 70.78\tiny{$\pm$2.79} & 63.26\tiny{$\pm$0.58} & 0.00\tiny{$\pm$0.00} & 1.37\tiny{$\pm$ 2.37} & 33.85 \tiny{$\pm$0.17}  \\ 
\rowcolor{lightgray} &  \textbf{CAMA-BC (Ours)} & A,T,V & \textbf{74.22}\tiny{$\pm$0.42} & \textbf{67.81}\tiny{$\pm$0.65} & \textbf{8.48}\tiny{$\pm$1.80} & \textbf{6.28}\tiny{$\pm$2.09}  & \textbf{39.20} \tiny{$\pm$0.14}  \\
\specialrule{1.2pt}{-1.2pt}{0pt}
\end{tabular}%
}
\vspace{-2mm}
\caption{ Evaluation of backchannel prediction on the KC-Dialog dataset and BACKSpeech dataset.} 
\label{tab:Main_BACKSpeech}
\vspace{-2mm}
\end{table*}


\noindent\textbf{Evaluation.}
We evaluate performance using the Macro F1 score, which is the average F1 score across all classes.
We report the best performance over 60 epochs with early stopping.
All results are averaged over three random seeds with statistical significance testing.

\noindent\textbf{Baselines.}
We compare our method with several baseline models utilizing different modality combinations of text, audio, and video. 
As text-only baselines, we use KoBERT \citep{skt2019kobert} and two large language models, GPT-4.1 \cite{gpt41} and Llama 4.0 Scout \cite{llama4}.
We adopt Ortega \citep{ortega2020oh} and BPM\_MT \citep{jang-etal-2021-bpm}, prior works on backchannel prediction that utilize both text and audio, as baselines in this work.
To examine whether visual information aids even in naive integration, we extend the existing Ortega and BPM\_MT models by incorporating a pre-trained VideoMAE \citep{tong2022videomae} via simple concatenation, resulting in Ortega-V and BPM-V. 
We also conduct experiments on GPT-4.1 \cite{gpt41} and Llama 4.0 Scout \cite{llama4} using text-only inputs, and on Gemini-2.5-Flash \cite{gemini25} for multimodal inputs, to verify the performance of backchannel prediction compared to well-known large language models.


\noindent\textbf{Implementation Details. }
Our model architecture incorporates pre-trained models specialized for each modality. 
For text processing, we utilize KoBERT \citep{skt2019kobert}, a Korean-specific variant of BERT \citep{Devlin2019BERTPO}, to effectively handle Korean language content.
Audio feature extraction utilizes HuBERT \citep{hsu2021hubert}, chosen for its self-supervised speech representations.
For video, we use VideoMAE \citep{tong2022videomae}, which is trained on Kinetics \citep{kay2017kinetics}, a dataset covering diverse human actions relevant to our task.

For the Ortega and BPM baselines, we follow the original feature settings reported in the respective papers, including their audio representations.

\subsection{Experimental Results}
Table \ref{tab:Main_BACKSpeech} demonstrates that CAMA-BC consistently outperforms baseline models across different architectures, achieving robust improvements in minority classes such as Continuer and Empathy, where contextual understanding is most critical.
A key finding emerges from comparing integration approaches. While direct visual feature concatenation (BPM-V) yields only modest improvement, as shown in Table \ref{tab:video_f1}, hierarchical alignment of the CAMA-BC achieves substantial gains.
This demonstrates that the value of visual information emerges through proper cross-modal alignment rather than naive feature combination.

Both large language models using text-only and multimodal inputs showed limited performance, confirming that backchannel prediction fundamentally requires multimodal reactive signals.
We observed substantial prompt sensitivity for the LLM/VLM baselines; however, despite exploring multiple prompt variants, the overall performance trend remained unchanged.

\subsection{Analysis}

\begin{table}[t]
\centering
\resizebox{\linewidth}{!}{%
\begin{tabular}{ccccccc}
\specialrule{1.2pt}{-1.2pt}{0pt}
\multirow{2}{*}{Model} & \multirow{2}{*}{NoBC} & \multicolumn{3}{c}{BC} & \multirow{2}{*}{Macro F1} \\ \cline{3-5}
 & & Continuer & Understanding  & Empathy & \\ \hline
CAMA-BC w/o MMA & 88.94 & 70.82 & 48.75 & 18.10 & 56.68 \\
CAMA-BC w/o MMA-BA & 88.98 & 69.71 & 49.64 & 17.96 & 56.57 \\
CAMA-BC w/o MMA-CA & 89.73 & 72.15 & 48.35 & 20.13 & 57.59 \\
\rowcolor{lightgray} \textbf{CAMA-BC (Ours)} & \textbf{89.90} & \textbf{72.73} & \textbf{49.99} & \textbf{21.49} & \textbf{58.53} \\
\specialrule{1.2pt}{-1.2pt}{0pt}
\end{tabular}%
}
\vspace{-2mm}
\caption{Evaluation on the impact of MMA-BA and MMA-CA on CAMA-BC on the KC-Dialog.}
\vspace{-3mm}
\label{tab:ablation_SelectStar}
\end{table}

\subsubsection{Ablation Study}
The detailed performance analysis in Table \ref{tab:Main_BACKSpeech} reveals that CAMA-BC with MMA components achieves powerful improvements in Continuer and Empathy classes, which represent minority classes in our dataset (see Table \ref{tab:sample}).
Table \ref{tab:ablation_SelectStar} shows the effect of MMA-CA and MMA-BA. 
The performance benefit persists even without MMA-BA, though with some degradation due to feature misalignment during transfer learning.

Interestingly, without MMA-BA, the model achieved only a moderate improvement in classification, resulting in the loss of reactive information and corruption of the representation space during the transfer process.
This suggests that enhancing the representation does not necessarily lead to improved classification.
On the other hand, without MMA-CA, the model achieves a better classification result because the enhanced knowledge directly aids the classification.
Our framework integrates both contextual understanding and reactive precision through MMA-CA and MMA-BA components.
The results show that the alignment loss serves as an unsupervised objective during Context Alignment and as a regularization during Backchannel Alignment, preventing catastrophic forgetting on transfer to classification.

\begin{table}[]
\resizebox{\columnwidth}{!}{%
\begin{tabular}{ccccc}
\specialrule{1.2pt}{-1.2pt}{0pt}
Model & Modality & NoBC & BC & Macro F1 \\ \hline
Llama 4.0 Scout & T & 28.80\tiny{$\pm$0.61} & 50.21\tiny{$\pm$0.34} & 39.50\tiny{$\pm$0.45} \\
GPT-4.1 & T & 44.15\tiny{$\pm$0.25} & 42.61\tiny{$\pm$0.41} & 43.38\tiny{$\pm$0.29} \\ \cdashline{1-5} 
Ortega & A,T & 68.27\tiny{$\pm$0.27} & 71.07\tiny{$\pm$1.04} & 69.67\tiny{$\pm$0.65} \\
BPM\_ST & A,T & 70.41\tiny{$\pm$0.52} & 77.55\tiny{$\pm$0.45} & 73.98\tiny{$\pm$0.45} \\ \cdashline{1-5}
Gemini-2.5-Flash & A,T,V & 67.24\tiny{$\pm$0.40} & 7.11\tiny{$\pm$0.74} & 37.18\tiny{$\pm$0.30} \\

Ortega-V & A,T,V & 70.27\tiny{$\pm$0.33} & 72.07\tiny{$\pm$0.80} & 71.17\tiny{$\pm$0.55} \\
BPM-V & A,T,V & 72.31\tiny{$\pm$0.48} & 78.34\tiny{$\pm$0.37} & 75.33\tiny{$\pm$0.39} \\
\rowcolor{lightgray} \textbf{CAMA-BC (Ours)} & A,T,V & \textbf{76.24}\tiny{$\pm$0.44} & \textbf{79.70}\tiny{$\pm$0.42} & \textbf{77.97}\tiny{$\pm$0.38} \\ 
 \specialrule{1.2pt}{-1.2pt}{0pt}
\end{tabular}%
}
\vspace{-2mm}
\caption{Evaluation on the performance of CAMA-BC on the SWBD Dataset with generated videos.}
\vspace{-4mm}
\label{tab:swbd}
\end{table}

\subsubsection{SWBD Dataset}
We experiment on the SWBD dataset to verify the effectiveness of CAMA-BC with the English dataset.
Since the original SWBD dataset lacks a visual modality, we generated video clips aligned with the corresponding audio segments, as described in Section~\ref{swbd_video}.
The experimental results are presented in Table~\ref{tab:swbd}.
Even with generated videos, visual modality consistently improves performance across all baselines.
CAMA-BC achieves the best overall performance among all models, demonstrating cross-linguistic generalizability.
These results suggest our approach captures the fundamental conversational dynamics rather than language-specific patterns.

\section{Conclusion}
We identified three critical limitations in natural interactive conversational AI: the systematic exclusion of visual information, temporal misalignment between modalities, and context-reaction imbalance.
Our Context-Aware Multimodal Alignment for Backchannel Prediction (CAMA-BC) represents the first systematic integration of visual information into backchannel prediction.
Through hierarchical cross-modal alignment, which addresses information asymmetry across modalities, CAMA-BC incorporates the visual modality without falling into the limitations of simple feature concatenation, which often fails to leverage the full potential of visual signals.
Comprehensive experiments across Korean (KC-Dialog, BACKSpeech) and English (SWBD) datasets demonstrated consistent improvements, suggesting that hierarchical cross-modal alignment not only enhances class-level prediction fidelity but also provides a language-agnostic framework that robustly captures universal conversational cues.

\section*{Limitations}
While our approach yields promising results, several limitations warrant discussion.
Although our modality hierarchy was empirically validated, it may not generalize well to different types of interaction.
This suggests the need for a dynamic hierarchy detection mechanism that adapts to varying conversational contexts.
Following prior work, we use a 1500ms audio window and a 5-word text context as a default.
The additional experiment in Table \ref{tab:ablation_longer} demonstrates that this may not fully capture long-range conversational dependencies.
Future work could explore adaptive or hierarchical management of temporal contexts.
Moreover, integrating video introduces latency and increased inference cost, potentially limiting real-time applicability; however, as detailed in Section \ref{sec:app:detail}, it is still affordable for real-time applications.
Finally, backchannel categorization inherently involves subjective properties. 
Our model inherits backchannel categories, such as "Empathy" or "Understanding," which are inherently subjective.
Table \ref{tab:ablation_noisy} demonstrates the robustness on the noisy labels, but it still implicitly inherits these biases without explicitly accounting for annotation uncertainty or inter-annotator disagreement.

\section*{Acknowledgments}
This research was supported by the “Advanced GPU Utilization Support Program” funded by the Government of the Republic of Korea (Ministry of Science and ICT), the National Research Foundation of Korea (NRF) grant funded by the Korea government (MSIT) (RS-2026-25480253), and Institute of Information \& communications Technology Planning \& Evaluation (IITP) grant funded by the Korea government (MSIT) (RS-2019-II190079, Artificial Intelligence Graduate School Program (Korea University), RS-2024-00457882, AI Research Hub Project, and RS-2024-00509257, Global AI Frontier Lab).

\bibliography{latex/custom}

\clearpage
\appendix
\nolinenumbers
\setcounter{equation}{0}
\renewcommand{\theequation}{A.\arabic{equation}}
\setcounter{table}{0}
\renewcommand{\thetable}{A.\arabic{table}}
\setcounter{figure}{0}
\renewcommand{\thefigure}{A.\arabic{figure}}
\setcounter{algorithm}{0}
\renewcommand{\thealgorithm}{A.\arabic{algorithm}}
\setcounter{page}{1}


\begin{table*}[t!]
\centering
\resizebox{0.9\textwidth}{!}{
\begin{tabular}{c|l}
\hline
\textbf{Symbol} & \textbf{Description} \\ \hline
$\mathcal{D}_\mathrm{a}$ & Audio dataset consisting of amplitude values sampled at rate $s_a$. \\
$\mathcal{D}_\mathrm{t}$ & An ordered set of words with their timestamps, where $p_i$ is a word and $t_i$ is its timestamp. \\
$\mathcal{D}_\mathrm{v}$ & Video dataset consisting of video frames sampled at rate $s_v$. \\
$\mathcal{D}_\mathrm{BC}$ & Backchannel dataset containing annotations for listener engagement events. \\
$D_{CA}$ & Context alignment dataset constructed from full dialogues for unsupervised context learning. \\
$\mathcal{W}^\mathrm{BC}_\mathrm{a}$, $\mathcal{W}^\mathrm{BC}_\mathrm{t}$, $\mathcal{W}^\mathrm{BC}_\mathrm{v}$ & Context windows for audio, text, and video modalities for backchannel prediction. \\
$\mathcal{W}^\text{CA}_\mathrm{a}$, $\mathcal{W}^\text{CA}_\mathrm{t}$, $\mathcal{W}^\text{CA}_\mathrm{v}$ & Context windows for audio, text, and video modalities for context alignment. \\
$\mathbf{a}_i, \mathbf{t}_i, \mathbf{v}_i$ & Audio, text, and video tensor for the $i$-th context window. \\
$\mathcal{B}_i$ & $i$-th batch of backchannel data, containing audio, text, and video tensors. \\
$\mathcal{J}$ & Set of selected layer groups $j=(j_a, j_t, j_v)$ used for multimodal alignment. \\
$\bar{\mathbf{a}}^{j_\mathrm{a}},\, \bar{\mathbf{t}}^{j_\mathrm{t}},\, \bar{\mathbf{v}}^{j_\mathrm{v}}$ & Average-pooled sample-level embeddings at selected layers $j_a, j_t, j_v$ for audio, text, and video. \\
$\bar{\mathbf{c}}_{vt}^{j},\, \bar{\mathbf{c}}_{va}^{j},\, \bar{\mathbf{c}}_{ta}^{j}$ & Average-pooled cross-attended representations of modality pairs at layer group $j$.\\
$\mathcal{S}_j$ & Set of average-pooled modality embeddings used for alignment at layer group $j$. \\
$W_\mathrm{a}$, $W_\mathrm{t}$, $W_\mathrm{v}$ & Linear transformations for audio, text, and video features. \\
$\mathcal{L}_\mathrm{MMA}^j$ & Multimodal Alignment loss at layer $j$. \\
$\mathcal{L}_\mathrm{MMA}^{j,*}$ & Multimodal Alignment loss with the cross-attention mechanism at layer group $j$. \\
$\mathcal{L}_\mathrm{CE}$ & Cross-entropy loss for backchannel classification. \\
$\delta(\mathbf{v},\mathbf{w})$ & Similarity function for aligning features $\mathbf{v}$ and $\mathbf{w}$. \\
$\mathbf{y}_{\mathrm{pred}}$ & Predicted logits for backchannel classification. \\
$W_h$ & Weight matrix for the final classification head. \\
$\mathbf{b}_h$ & Bias vector for the final classification head. \\
$\lambda$ & Hyperparameter controlling the contribution of the alignment loss. \\
$i_\mathrm{a}$, $i_\mathrm{t}$, $i_\mathrm{v}$ & Number of layers in audio, text, and video encoders. \\
$n_l$ & Number of layers used for alignment in the Multimodal Alignment framework. \\
\hline  
\end{tabular}
}
\caption{ Description for the mathematical symbols used in the paper.}
\label{tab:math_notation}
\vspace{-2mm}

\end{table*}

\section{Mathematical Notations}
Table \ref{tab:math_notation} provides a comprehensive reference for the mathematical symbols and notations used throughout the paper.

\section{Additional Experimental Results}
\subsection{Ablations for Hyperparameters}
We conduct experiments to evaluate the effectiveness of the components that comprise our method.
First, in Table \ref{tab:ablation_lambda}, we see the results of adjusting the lambda scale constant in our method.
Overall, we observe that it performs nearly as well as if it is not too large, so it does not violate the classification entropy loss.
In Table \ref{tab:ablation_depth}, we also observed that selecting more layers towards the front of the model did not guarantee better performance when aligning layers.
This is likely because, as discussed in Section \ref{sec:mma}, the early, abstracted information hinders alignment.

\subsection{Ablation for Alignment Methods}

\begin{table}[th!]
\centering
\resizebox{\linewidth}{!}{%
\begin{tabular}{cccccccc}
\specialrule{1.2pt}{-1.2pt}{0pt}
\multirow{2}{*}{$\lambda$} & \multirow{2}{*}{NoBC} & \multicolumn{3}{c}{BC} & \multirow{2}{*}{Macro F1}  \\ \cline{3-5}
 & & Continuer & Understanding  & Empathy & \\ \hline
1.0 & 89.57 & 71.75 & 46.63 & 14.87 & 55.71 \\
0.5 & 89.68 & 72.68 & 47.47 & 15.80 & 56.41 \\
0.2 & 90.02 & 72.81 & 49.07 & 18.98 & 57.72 \\
\rowcolor{lightgray} \textbf{0.1 (Ours)} & 89.90 & 72.73 & 49.99 & 21.49 & \textbf{58.53} \\
0.05 & 89.68 & 71.71 & 47.60 & 21.74 & 57.68 \\
0 & 88.98 & 69.71 & 49.64 & 17.96 & 56.57 \\
\specialrule{1.2pt}{-1.2pt}{0pt}
\end{tabular}%
}
\vspace{-2mm}
\caption{Ablation study on the loss weight {$\lambda$} on the KC-Dialog dataset.}
\label{tab:ablation_lambda}
\end{table}
\begin{table}[th!]
\centering
\resizebox{\linewidth}{!}{%
\begin{tabular}{cccccccccc}
\specialrule{1.2pt}{-1.2pt}{0pt}
 & &  & \multicolumn{3}{c}{BC} &  \\ \cline{4-6}
\multirow{-2}{*}{$n_l$} & \multirow{-2}{*}{Layer} & \multirow{-2}{*}{NoBC} & Continuer & Understanding  & Empathy & \multirow{-2}{*}{Macro F1}  \\ \hline
11 & 0-11 & 90.07 & 73.07 & 48.11 & 21.16 & 58.10 \\
6 & 6-11 & 90.05 & 73.00 & 49.63 & 20.21 & 58.22 \\
\rowcolor{lightgray}
3 & 9-11 & 89.90 & 72.73 & 49.99 & 21.49 & \textbf{58.53}  \\
1 & 11 & 89.73 & 72.11 & 48.40 & 22.59 & 58.21  \\
\specialrule{1.2pt}{-1.2pt}{0pt}
\end{tabular}%
}
\caption{Analysis of the impact of alignment depth in Multi-Layer Multimodal Alignment (MMA) on the KC-Dialog dataset.}
\label{tab:ablation_depth}
\end{table}
\begin{table}[th!]
\centering
\resizebox{\linewidth}{!}{%
\begin{tabular}{ccccccc}
\specialrule{1.2pt}{-1.2pt}{0pt}
\multirow{2}{*}{Alignment} & \multirow{2}{*}{NoBC} & \multicolumn{3}{c}{BC} & \multirow{2}{*}{Macro F1} \\ \cline{3-5}
 & & Continuer & Understanding  & Empathy & \\ \hline
Praveen & 88.17 & 69.69 & 49.85 & 19.33 & 56.75 \\ 
Sun & 89.46 & 71.84 & 43.59 & 22.26 & 56.79 \\
\rowcolor{lightgray} \textbf{MMA} & 89.90 & 72.73 & 49.99 & 21.49 & \textbf{58.53} \\ 
\specialrule{1.2pt}{-1.2pt}{0pt}
\end{tabular}%
}
\caption{Comparison of Multi-Layer Multimodal Alignment (MMA) with other alignment methods on the KC-Dialog dataset.}
\vspace{-2mm}
\label{tab:ablation_other_alignment}
\end{table}
\begin{table*}[th]
\centering
\resizebox{0.9\textwidth}{!}{%
\begin{tabular}{ccccccccccc}
\specialrule{1.2pt}{-1.2pt}{0pt}
 & &  & \multicolumn{3}{c}{BC} &  \\ \cline{5-7}
\multirow{-2}{*}{Model} & \multirow{-2}{*}{Text Length} & \multirow{-2}{*}{Audio/Video Length} & \multirow{-2}{*}{NoBC} & Continuer & Understanding  & Empathy & \multirow{-2}{*}{Macro F1}  \\ \hline
 & 5 words & 1500ms & 86.96\tiny{$\pm$0.18} & 65.72\tiny{$\pm$1.49} & 45.50\tiny{$\pm$1.97} & 16.89\tiny{$\pm$2.09} & 53.77\tiny{$\pm$0.09} \\
\multirow{-2}{*}{BPM-V} & 10 words & 3000ms & 88.99\tiny{$\pm$0.37} & 68.01\tiny{$\pm$0.33} & 46.92\tiny{$\pm$0.76} & 17.01\tiny{$\pm$0.15} & 55.50\tiny{$\pm$0.30} \\ \cdashline{1-8}
 & 5 words & 1500ms & 89.90\tiny{$\pm$0.07} & 72.73\tiny{$\pm$0.13} & 49.99\tiny{$\pm$0.25} & 21.49\tiny{$\pm$0.49} & 58.53\tiny{$\pm$0.07}  \\
\multirow{-2}{*}{\textbf{CAMA-BC (Ours)}} & 10 words & 3000ms & 91.21\tiny{$\pm$0.37} & 73.91\tiny{$\pm$0.23} & 51.14\tiny{$\pm$0.22} & 22.69\tiny{$\pm$0.12} & 59.76\tiny{$\pm$0.11}  \\
\specialrule{1.2pt}{-1.2pt}{0pt}
\end{tabular}%
}
\caption{Analysis of backchannel prediction performance with different input lengths on the KC-Dialog dataset.}
\label{tab:ablation_longer}
\end{table*}
\begin{table}[th!]
\centering
\resizebox{\linewidth}{!}{%
\begin{tabular}{cccccccccc}
\specialrule{1.2pt}{-1.2pt}{0pt}
 & &  & \multicolumn{3}{c}{BC} &  \\ \cline{4-6}
\multirow{-2}{*}{Model} & \multirow{-2}{*}{Ratio} & \multirow{-2}{*}{NoBC} & Continuer & Understanding  & Empathy & \multirow{-2}{*}{Macro F1}  \\ \hline
 & 0\% & 86.96 & 65.72 & 45.50 & 16.89 & 53.77 \\
 & 5\% & 86.73 & 64.87 & 0.00 & 0.00 & 37.90 (-15.87) \\
\multirow{-3}{*}{BPM-V} & 10\% & 86.46 & 64.55 & 0.00 & 0.00 & 37.75 (-16.02) \\ \cdashline{1-7}
 & 0\% & 89.90 & 72.73 & 49.99 & 21.49 & 58.53  \\
 & 5\% & 88.25 & 69.92 & 44.96 & 11.81 & 53.74 (-4.79) \\
\multirow{-3}{*}{\textbf{CAMA-BC}} & 10\% & 87.99 & 69.55 & 43.08 & 10.04 & 52.67 (-5.86) \\
\specialrule{1.2pt}{-1.2pt}{0pt}
\end{tabular}%
}
\caption{Performance of BPM-V and CAMA-BC under different ratios of synthetic label noise on the KC-Dialog dataset, where a fixed percentage  of training labels was randomly flipped.}
\label{tab:ablation_noisy}
\end{table}

We checked the performance impact on the backchannel prediction task for different multimodality alignment-seeking methodologies.
In many cases, techniques such as cross-attention are designed to use two alignment targets, which creates structural problems when applied to our scenario, where three modality alignments are required.
As shown in Table \ref{tab:ablation_other_alignment}, when we expand these methodologies, they perform relatively poorly, demonstrating the importance of the assistance and selectivity between modalities that we sought.

\subsection{Analysis of Input Length}

We conducted experiments to evaluate the impact of input length on model performance.
Specifically, as shown in Table \ref{tab:ablation_longer}, we compared a 5-word, 1500ms input with an extended 10-word, 3000ms input and observed that increasing the input length provides additional contextual cues, leading to modest performance improvements.
However, our primary goal is not to maximize performance by simply expanding the input window.
Instead, we focus on designing efficient and generalizable architectures that perform reliably under practical constraints, ensuring robust performance even with limited input length.

\subsection{Robustness to Annotation Noise}

Backchannel annotations are created according to detailed guidelines to maintain consistency; however, there is no absolute ground truth, and subjectivity is inevitably involved.
To examine how such annotation disagreement and noise affect model robustness, we conducted a synthetic label noise experiment.
Specifically, we randomly flipped a portion of the training labels (5\% and 10\%) to simulate annotator disagreement and evaluated model performance under these noisy conditions.
The results, shown in Table \ref{tab:ablation_noisy}, indicate that BPM-V suffers substantial drops in Macro F1 as the noise ratio increases, especially for minority classes.
In contrast, CAMA-BC shows only a modest performance decline and remains stable even when label noise is introduced.
This suggests that our model is not heavily affected by label quality uncertainty and learns stable multimodal representations, maintaining robustness even under subjective labeling noise.

\subsection{Qualitative Results}

\begin{figure*}
    \includegraphics[width=\linewidth]{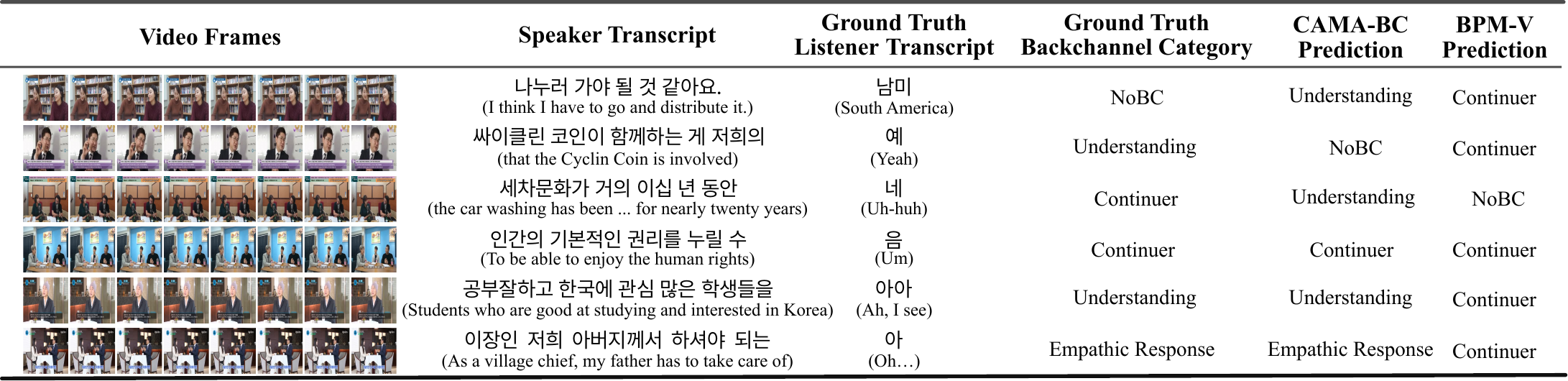}
    \caption{The samples of the backchannel prediction in the BACKSpeech dataset, ground truth backchannel category, and the prediction of CAMA-BC and BPM-V.}
    \label{fig:sample}
\end{figure*}

\begin{figure*}[h!]
    \includegraphics[width=\linewidth]{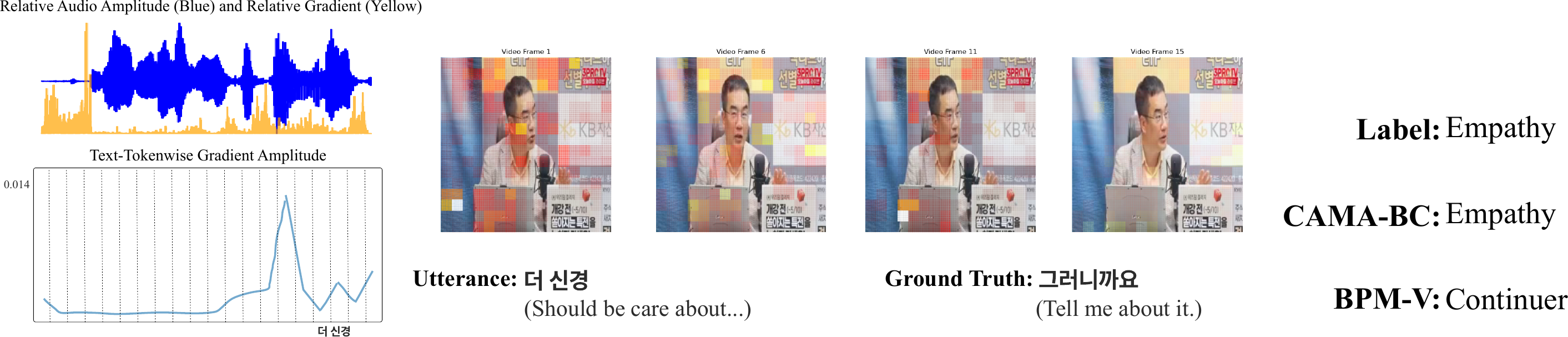}
    \caption{Gradient visualization for successful sample in CAMA-BC.}
    \label{fig:succ}
\end{figure*}

\begin{figure*}[h!]
    \includegraphics[width=\linewidth]{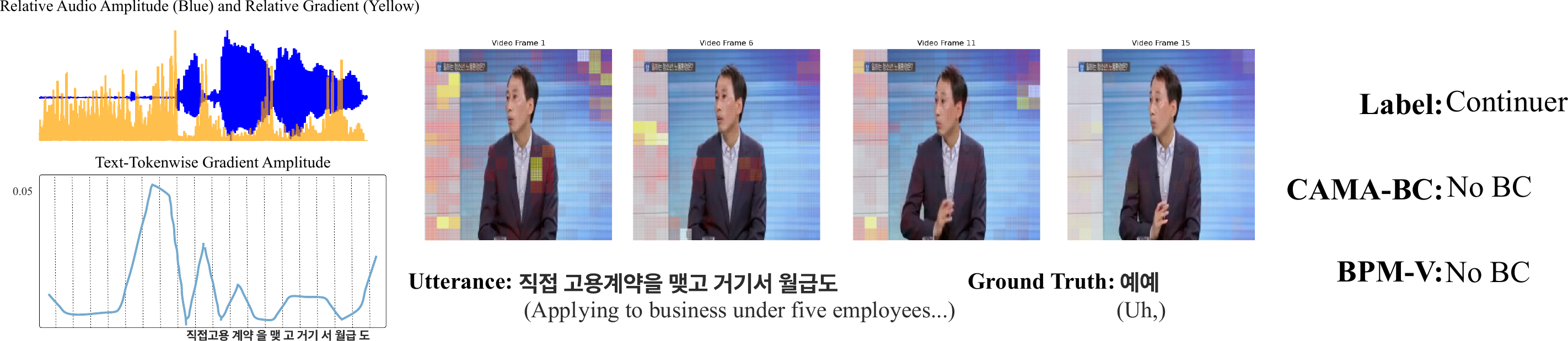}
    \caption{Gradient visualization for failure sample in CAMA-BC.}
    \label{fig:fail}
\end{figure*}

Figure \ref{fig:sample} shows prediction examples, while Figures \ref{fig:succ} and \ref{fig:fail} visualize attention patterns in successful and failed CAMA-BC cases to identify the most essential multimodal features.
The visual modality analysis reveals consistent attention to early video frames, relatively capturing more pre-utterance cues than other modalities.
Additionally, failure cases indicate that the audio and text are not informative enough, and the speaker is using minimal gestures and expressions.
This demonstrates misaligned visual attention where the model inadequately focuses on speaker-related content.
In both cases, the audio tends to focus on the silence between utterances, which is a significant indicator of interaction.




\section{SWBD Video Dataset}

\subsection{SWBD Dataset}
\label{swbd}
We utilize the Switchboard (SWBD) corpus to evaluate the effectiveness of CAMA-BC in another language setting.
SWBD is a large-scale English telephone speech dataset comprising over 2,400 conversations with more than 500 adult speakers discussing a wide range of topics.
Each conversation lasts approximately 6–10 minutes and includes detailed utterance-level transcriptions for all audio recordings.

For text and audio feature extraction, we analyze a temporal window of $n = 1500 \text{ms}$ and $m = 5$ words following each backchannel onset, consistent with our previous experiments.
As SWBD does not contain video, we generate video data synthetically for the same temporal segments to incorporate the visual modality.

\begin{table}[t]
\centering
\resizebox{1.\linewidth}{!}{%
\begin{tabular}{ccc||cccc||c}
\specialrule{1.2pt}{-1.2pt}{0pt}
Dataset & NoBC & BC & Train & Validation & Test & Total & \#Samples for MMA-CA \\ \hline
SWBD & 48,087 & 48,087 & 76,938 & 4,808 & 14,428 & 96,174 & 194,178 \\
\specialrule{1.2pt}{-1.2pt}{0pt}
\end{tabular}%
}
\caption{Distribution of each backchannel category and the number of samples used for MMA-CA. Following prior works, we ensured that the NoBC and BC samples had equal numbers. The dataset was then split into training, validation, and test sets in an $ 8:0.5:1.5$ ratio.}
\label{tab:sample_swbd}
\end{table}

\begin{figure}[h!]
    \centering
      \includegraphics[width=\linewidth]{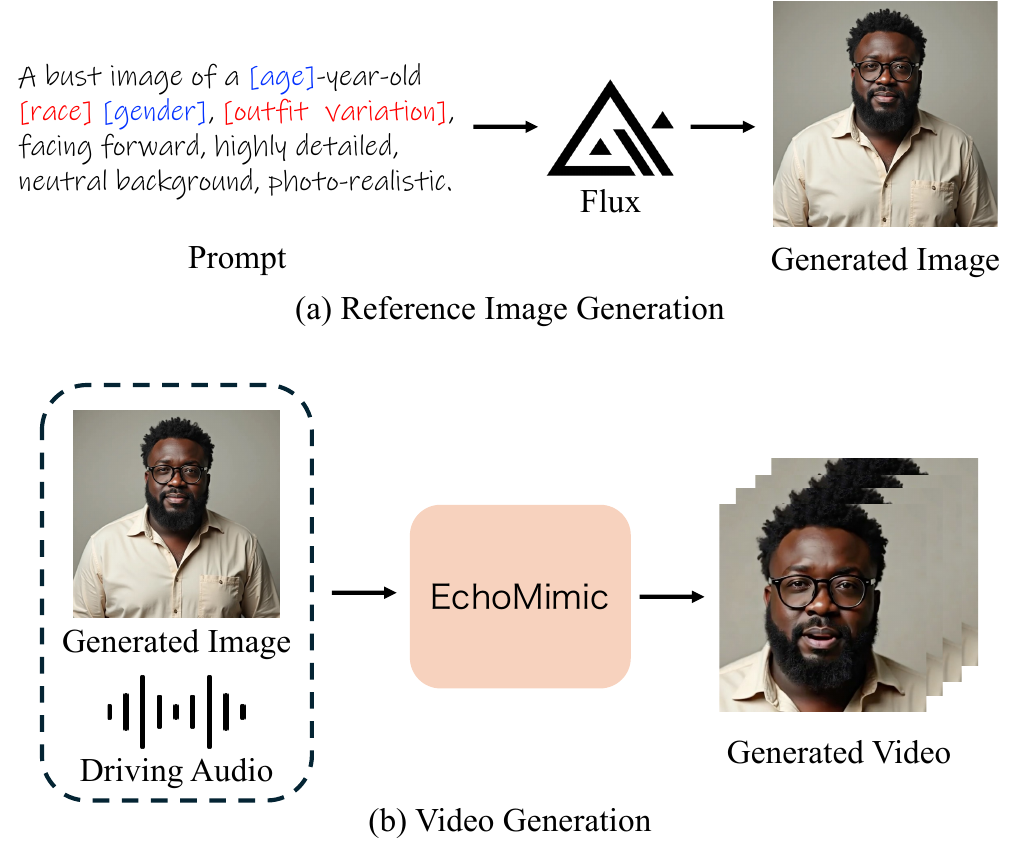}
    \caption{Image–video generation pipeline: (a) Reference bust image generation with Flux.1 [dev] \cite{flux2024} using diverse prompts, and (b) video generation guided by EchoMimic \citep{chen2025echomimic} with driving audio input.}
    \label{fig:genpipeline}
\end{figure}

\subsection{Video Generation}
\label{swbd_video}
\subsubsection{Reference Image Generation}
To address the absence of visual data in the SWBD corpus, we first generate reference images for each speaker.
We utilized the Flux.1 dev \cite{flux2024} model to synthesize high-quality bust images conditioned on speaker metadata.
The generation prompt is defined as:
\textit{"A bust image of a \texttt{[age]}-year-old \texttt{[race]} \texttt{[gender]}, \texttt{[outfit variation]}, facing  forward, highly detailed, neutral background, photo-realistic."}

\begin{figure*}[h!]
    \centering
      \includegraphics[width=1.0\textwidth]{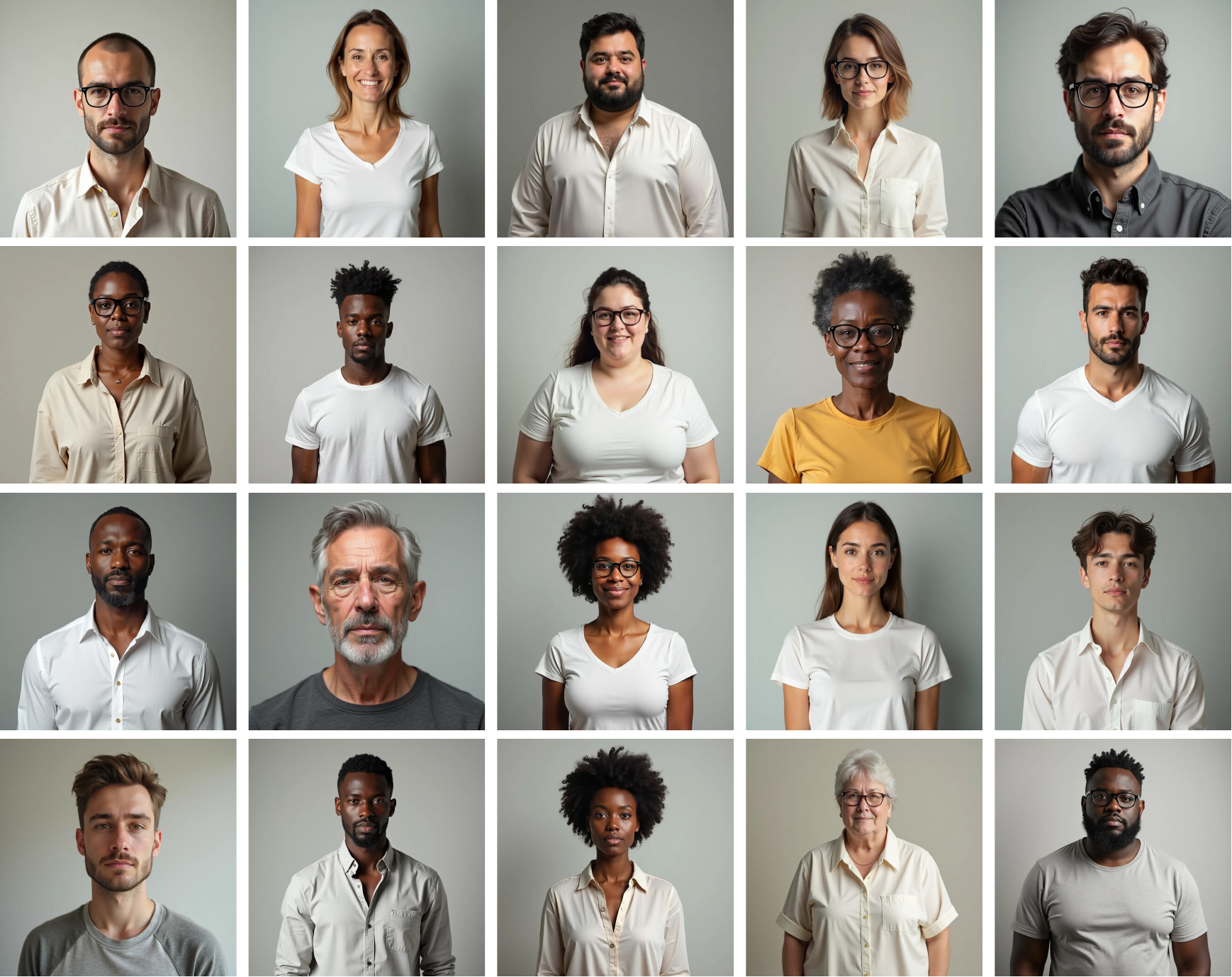}
    \caption{Reference image generation results with Flux.1 [dev] \cite{flux2024} using diverse prompts for age, gender, race and outfit variations.}
    \label{fig:genimg}
\end{figure*}

Here, \texttt{[age]} and \texttt{[gender]} are extracted directly from the SWBD metadata.
At the same time \texttt{[race]} and \texttt{[outfit variation]} are randomly selected from predefined lists — \textit{ \texttt{[“}White\texttt{”, “}Middle Eastern\texttt{”, “}Black\texttt{”, “}Mediterranean\texttt{”]}} for race and \textit{ \texttt{[“}wearing clothes\texttt{”, “}wearing shirts\texttt{”, “}wearing t-shirts\texttt{”]}} for outfit — to ensure visual diversity.
This process is illustrated in Figure~\ref{fig:genpipeline} (a).
An example of the generated reference images is shown in Figure~\ref{fig:genimg}.
Flux.1 dev achieved PDist of 0.332, SSIM of 0.896, and PSNR of 31.1, demonstrating its ability to generate high-quality, photorealistic images and supporting its suitability for our reference image generation stage.

\begin{figure*}[h!]
    \centering
      \includegraphics[width=1.0\textwidth]{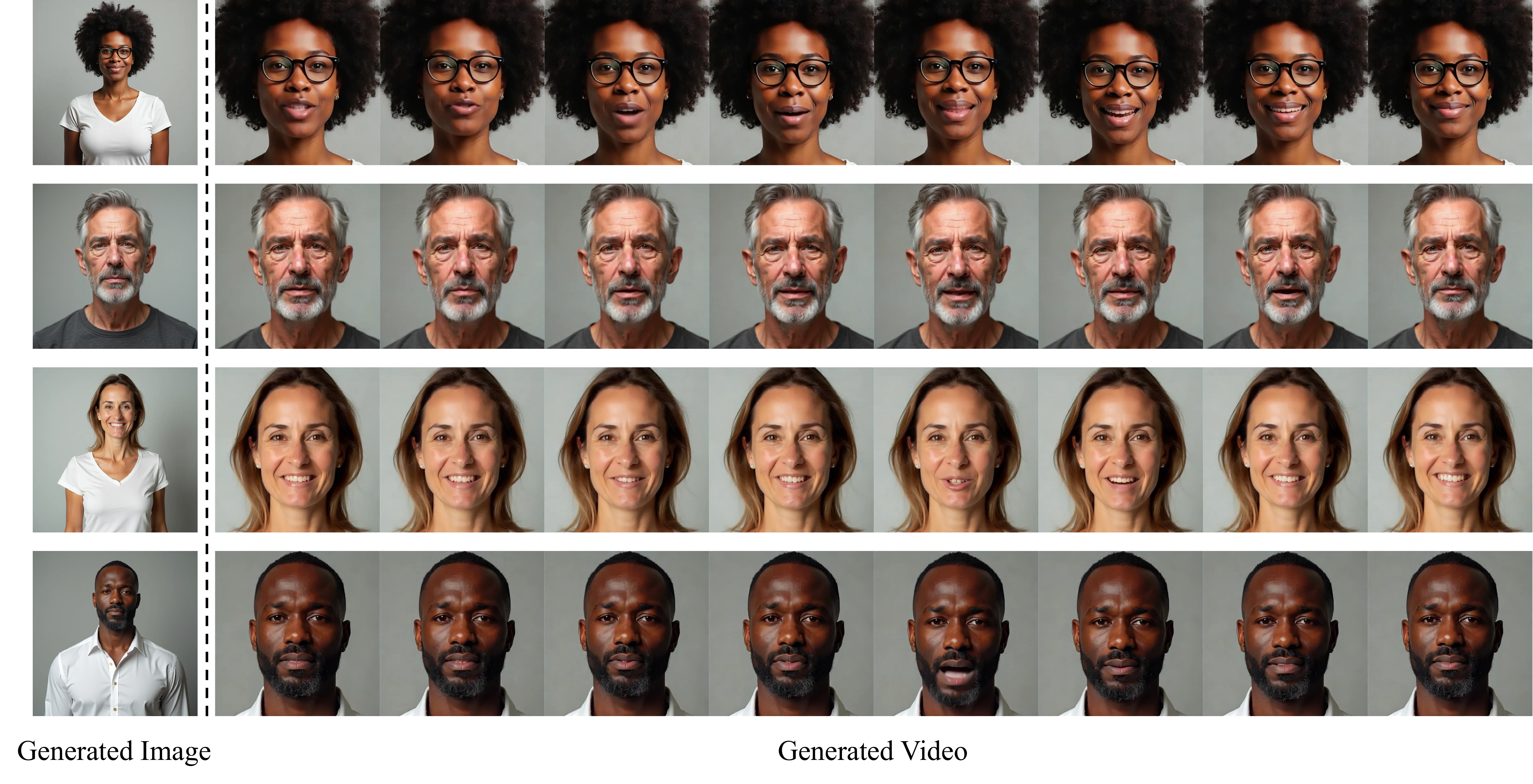}
    \caption{Video generation results guided by EchoMimic \citep{chen2025echomimic} using the driving audio and reference images.}
    \label{fig:genvid}
\end{figure*}

\subsubsection{Video Generation}
Based on the generated reference images, we synthesize videos for the SWBD corpus to incorporate visual modality information.
Aligned with the audio segments, we generate a video clip for each backchannel occurrence covering a temporal window of $n = 1500 \text{ms}$ following the backchannel onset.
To produce realistic lip movements and facial expressions synchronized with the speech, we use Echomimic, which takes the SWBD audio as input and animates the reference image accordingly.
This approach compensates for the absence of original video in SWBD and ensures consistency with the multimodal input setting used for the other datasets.
The overall video generation process is illustrated in Figure~\ref{fig:genpipeline} (b), and examples of the generated video frames are demonstrated in Figure~\ref{fig:genvid}.
Echomimic achieved an FID of 29.14, FVD of 492.78, SSIM of 0.812, and E-FID of 1.11, demonstrating its ability to generate temporally consistent and visually coherent videos, which supports its suitability for our video synthesis stage.

\section{Additional Implementational Details}
\label{sec:app:detail}

In CAMA-BC, MMA is applied at layers $\mathcal{J}=\{9, 10, 11\}$.
We apply different learning rates across model components: $5\times10^{-6}$ for encoder fine-tuning, $5\times10^{-5}$ for classifier training, and $5\times10^{-4}$ for randomly initialized projection and cross-attention layers.
A per-epoch learning rate decay of 0.95 is applied to the encoder, projection, and cross-attention layers. 
All experiments were conducted using 4 NVIDIA RTX 3090 GPUs with a fixed batch size of 16.
In the same computational environment, CAMA-BC requires $20.083\text{ ms}$, while BPM-V requires $18.868\text{ ms}$, supporting the efficiency of backchannel prediction applications.
The reported inference time is measured end-to-end per sample, including feature extraction from all modalities, without feature caching.

We evaluated GPT-4.1 and Llama 4.0 Scout using carefully structured prompts that included comprehensive task descriptions and standardized response formats.
To ensure systematic analysis, the models were instructed to provide explicit reasoning and confidence scores alongside their backchannel predictions, allowing for a detailed examination of their decision-making processes.
The complete prompt structure for Korean datasets (KC-Dialog and BACKSpeech) is provided in \ref{prompt:KCnBACK}.
Since SWBD uses a binary backchannel classification scheme rather than the four-class system employed in Korean datasets, we adapted the prompts accordingly, as shown in Appendix \ref{prompt:SWBD}.
They showed lower performance and sensitivity to prompts for both KC-Dialog (23.71 and 26.75) and BACKSpeech (13.34 and 19.89), so we reported the Text modality as a baseline.
We also evaluated a Gemini-2.5 Flash as a Multimodal LLM.
For the Gemini-2.5 Flash, we utilized the same format of instructions, adding a video and audio input in 
\texttt{ \{example\_transcript\} } and \texttt{ \{request\_transcript\} } after the transcript; the format of multimodal input of Gemini is not publicly available.

\onecolumn
\paragraph{Prompt 1} Text prompt for Llama 4.0 Scout and GPT-4.1 on KC-Dialog and BACKSpeech dataset.
\label{prompt:KCnBACK}
\begin{quote}
\begin{lstlisting}[style=PromptStyle]
# Korean Conversation Backchannel Analyzer
You will analyze short segments (max 5 words) of Korean conversations and classify the response type.
Following the instructions below.


## Definition
A backchannel is a brief response that indicates listener engagement without adding direct content to the conversation.

## Response Types
- **NoBC**: No backchanneling occurs
- **Continuer**: Encourages speaker to continue
- **Understanding**: Shows comprehension
- **Empathic**: Conveys emotional reaction

## Instructions
1. Analyze the given dialogue context
2. Select the most appropriate response type
3. Return your answer in JSON format:
```json
{
  "response_type": "[NoBC|Continuer|Understanding|Empathic]",
  "confidence": 0-1,
  "reasoning": "Brief explanation"
}
```

## Examples

### class: {example_BC_category}
Context: {example_transcript}
Response: {ground_truth}

... (n-shot for each class) ...

Now, analyze the following dialogue segment and classify the response type.
Context: {request_transcript}
\end{lstlisting}
\end{quote}
\twocolumn

\onecolumn
\paragraph{Prompt 2} Text prompt for Llama 4.0 Scout and GPT-4.1 on SWBD dataset.
\label{prompt:SWBD}
\begin{quote}
\begin{lstlisting}[style=PromptStyle]
# Backchannel Analyzer
You will analyze short segments (max 5 words) of conversations and classify
if the response is a backchannel or not.
Following the instructions below.

## Definition
A backchannel is a brief response that indicates listener engagement
without adding direct content to the conversation.

## Response Types
- **NoBC**: No backchanneling occurs
- **BC**: Backchanneling occurs

## Instructions
1. Analyze the given dialogue context
2. Select the most appropriate response type
3. Return your answer in JSON format:
```json
{
  "response_type": "[NoBC|BC]",
  "confidence": 0-1,
  "reasoning": "Brief explanation"
}
```
## Examples

### class: {example_BC_category}
Context: {example_transcript}
Response: {ground_truth}

... (n-shot for each class) ...

Now, analyze the following dialogue segment and classify the response type.
Context: {request_transcript}
\end{lstlisting}
\end{quote}
\twocolumn

\end{document}